# Organ at Risk Segmentation in Head and Neck CT Images Using a Two-Stage Segmentation Framework Based on 3D U-Net

**Yueyue Wang[1,2], Liang Zhao[1,2], Manning Wang[1,2], and Zhijian Song[1,2]**
[1]Digital Medical Research Center, School of Basic Medical Sciences, Fudan University, Shanghai, China
[2]Shanghai Key Laboratory of Medical Imaging Computing and Computer Assisted Intervention, Shanghai, China

Corresponding author: Manning Wang (e-mail: mnwang@fudan.edu.cn) and Zhijian Song (e-mail: zjsong@fudan.edu.cn)

**ABSTRACT** Accurate segmentation of organs at risk (OARs) plays a critical role in the treatment planning of image-guided radiotherapy of head and neck cancer. This segmentation task is challenging for both humans and automated algorithms because of the relatively large number of OARs to be segmented, the large variability in size and morphology across different OARs, and the low contrast between some OARs and the background. In this study, we propose a two-stage segmentation framework based on 3D U-Net. In this framework, the segmentation of each OAR is decomposed into two subtasks: locating a bounding box of the OAR and segmenting the OAR from a small volume within the bounding box, and each subtask is fulfilled by a dedicated 3D U-Net. The decomposition makes each subtask much easier so that it can be better completed. We evaluated the proposed method and compared it to state-of-the-art methods using the Medical Image Computing and Computer-Assisted Intervention 2015 Challenge dataset. In terms of the boundary-based metric 95% Hausdorff distance, the proposed method ranked first for seven of nine OARs and ranked second for the other OARs. In terms of the area-based metric dice similarity coefficient, the proposed method ranked first for five of nine OARs and ranked second for the other three OARs with a small difference from the method that ranked first.

## I. INTRODUCTION

Head and neck (HaN) cancer is one of the most common cancers, with more than half a million cases worldwide per year [1]. Image-guided radiation therapy (IGRT), including intensity-modulated radiation therapy (IMRT) and volumetric modulated arc therapy, is a state-of-the-art treatment option because of its highly conformal dose delivery [2]–[4]. The key to the success of IGRT is patient-specific treatment planning, in which medical images are used to make a radiation plan to concentrate the radiation dose on the target volume while minimizing the dose to the surrounding organs at risk (OARs). Therefore, it is essential to segment the OARs in treatment planning images, which usually include HaN computed tomography (CT) images. In current clinical practice, OARs are usually delineated manually, but the complexity and variability of the OARs morphology in HaN CT images make it an inaccurate and very time consuming task [5], [6]. It may take radiologist three hours to segment all OARs for treatment planning [5]. Some treatment planning systems have automatic segmentation function, such as the atlas-based segmentation methods [7], but the segmentation result has not met the clinical needs. Intensive labor is still needed for manual adjustment of the segmentation result to make it applicable for treatment planning and the time needed for manual adjustment is comparable to manual segmentation from scratch [6]. Therefore, there is a great demand for a rapid, accurate, and automatic OAR segmentation method to reduce radiologist labor in HaN treatment planning.

Medical image segmentation is an area of intense research, and many methods for segmenting different targets from medical images of different modalities have been proposed. Some of these methods have also been applied in OAR segmentation, but unfortunately, the current results are far from being satisfactory. A Head and Neck Auto Segmentation Challenge was held in conjunction with the Medical Image Computing and Computer-Assisted Intervention (MICCAI) conference in 2015 (referred to as the "MICCAI 2015 Challenge" from here on), which provided a public data set for OARs segmentation in HaN CT images [8]. Six teams participated in this challenge and finished this task using different segmentation methods, including the statistical shape model, active appearance model, multiatlas-based segmentation method and the semiautomatic segmentation method [8], but their segmentation results were not satisfactory to radiologist. The challenges of OAR segmentation in HaN CT images include: (i) the complexity and variability of the OARs are high, and it is difficult to incorporate prior information into shape models to support the segmentation of new images; (ii) the sizes of OARs are varied, and most segmentation methods usually get accurate results in bigger OARs while inaccurate results in smaller OARs and (iii) the contrast of soft tissues is poor in CT images, which makes it difficult to segment some OARs, such as brainstem.

Although the contrast between bone and soft tissues is relatively high in CT images, the characteristics of the HaN OAR segmentation task, including the large number of OARs to be segmented, the great variety in size and morphology of different OARs, and the low contrast between some OARs and their background, make simple segmentation methods, such as thresholding, edge detection, and region growing, difficult to succeed. Many methods that have been successfully used in other medical image segmentation tasks, such as 3D level set [9] and atlas-based techniques [10], have also been applied in this field, but the results are not satisfactory.

Several approaches have been developed to incorporate prior knowledge, which often represents the results of gold standard segmentation of some subjects, to help segment new subjects, and these approaches have also been used in HaN OAR segmentation. For example, the method proposed in [11] built a statistical shape model of OARs and deforms the model to fit the image to achieve segmentation. A multiatlas approach [12] registered the segmented images to the target image and then fused the label of the segmented images to obtain a segmentation result of the target image. Another approach is to train a classifier with prior segmented images and transform the segmentation task into a classification task [13]. In the MICCAI 2015 Challenge, most teams adopted several approaches that include the statistical shape model, active appearance model, and the multiatlas-based method to utilize prior knowledge. This challenge provided a unified evaluation framework for different methods on OAR segmentation.

In recent years, deep learning methods, especially the convolutional neural network (CNN), have demonstrated excellent performance in medical image segmentation tasks [14]–[19], and CNN has also been applied for OARs segmentation in H&N CT images[20]–[23]. The first study [20] using deep learning methods proposed a 2D CNN for OARs segmentation from in-house HaN CT images, but it only got a slight improvement in right submandibular gland and right optic nerve, and the performance for the other OARs was similar to that of the traditional methods. In [21], a interleaved 3D CNNs method was proposed to jointly segment the optic nerve and chiasm. They used atlas-based method to locate a bounding box enclosing the target OAR and then performed segmentation in a small target volume. Zhu et al. [22] proposed the AnatomyNet, an end-to-end and atlas-free three dimensional squeeze-and-excitation U-Net (3D SE U-Net), for fast and fully automated whole-volume HaN anatomical segmentation. Tong et al. [23] proposed a fully convolutional neural network with a shape representation model for multi-organ segmentation for HaN cancer radiotherapy. However, these existing deep-learning-based methods generally produced accurate segmentation maps for large organs, while the accuracy of small OARs was often sacrificed.

To seperate the segmentation of large and small OARs, we adopt a two-stage framework for OARs localization and segmentation. Recently, two-stage framework and U-Net have shown their outstanding performances in various medical image computing tasks [24]–[30]. In this study, we propose a two-stage framework to decompose OAR segmentation into two relatively simpler tasks and complete each task by a dedicated 3D U-Net. The first task is to locate the target OAR with a bounding box and the second task is to segment the target OAR within the bounding box. Decomposition of this task makes it simpler than directly segmenting the OARs from the entire volume and improves the segmentation performance. Experiments using MICCAI 2015 Challenge data showed that the proposed method achieved the highest dice similarity coefficient (DSC) for six of the nine OARs and achieved the second highest DSC for the other three OARs. In addition, the proposed method achieved the smallest 95% Hausdorff distance (95HD) for seven of the nine OARs with a significant benefit and achieved the second smallest 95HD for the other two OARs.

## II. MATERIALS AND METHODS

### A. THE MICCAI 2015 CHALLENGE DATASET

In this study, we evaluated the proposed OAR segmentation framework and compared it to other methods using the PDDCA dataset, which is publicly available at (http://www.imagenglab.com/newsite/pddca/). This dataset was provided by Dr. Gregory C. Sharp and was used in the Head and Neck Auto-Segmentation Challenge 2015, a satellite event at the MICCAI 2015 conference. The current version (v 1.4.1) of the PDDCA dataset consists of 25 training images, 8 additional training images, and 15 testing images. The original images are from the RTOG 0522 clinical trial [18], which provides 111 HaN CT images for treatment planning. The subset was chosen to ensure that the image quality is adequate and the target OARs have minimal overlap with the tumors. Each image consists of a series of axial slices with 512 × 512 voxels on each slice, and the number of slices varies from 76 to 263. The in-plane spacing is between 0.76 mm × 0.76 mm and 1.27 mm × 1.27 mm, and the inter-plane spacing is between 1.25 and 3.00 mm.

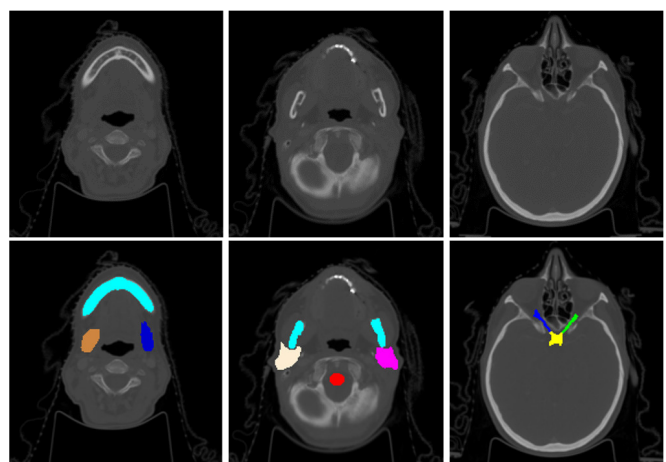

Mandible, Submandibular left, Submandibular right, Parotid left, Parotid right, Brainstem, Optic nerve left, Optic nerve right, Chiasm

**FIGURE 1.** Examples of OARs of a patient in different slices of a CT scan, and the OARs are manually annotated and shown in different colors.

In this dataset, nine anatomical structures, namely, the brainstem, optic nerve left, optic nerve right, chiasm, parotid left, parotid right, mandible, submandibular left, and submandibular right, were used as segmentation targets. And examples of OARs of a patient in different slices of a CT scan are shown in Fig. 1. All nine of these structures are important OARs in HaN radiotherapy [19] and they are manually segmented by experts to provide high quality and consistency. The mask for most of these structures are provided in all 33 training images, except that the mandibular, left submandibular glands and right submandibular glands are only segmented in 25, 26 and 21 training images, respectively. The masks for all nine structures are provided in 15 testing images and used as the gold standard for evaluation.

### B. OVERVIEW OF THE TWO-STAGE SEGMENTATION FRAMEWORK

The proposed two-stage segmentation framework and its training and testing flowcharts are illustrated in Fig. 2. The framework consists of two 3D U-Nets. The original images and masks were first cropped to a volume with a consistent resolution of 384 × 384 × 224 for further processing.

The first 3D U-Net, denoted as LocNet, is used to coarsely locate the target structure with a bounding box. The cropped images and masks are first downsampled to a resolution of 96 × 96 × 56 in voxels and used for training LocNet. LocNet outputs a 0–1 classification for each voxel, indicating whether a voxel falls in the bounding box. A postprocessing step is used to generate a bounding box of size (h/4) × (w/4) × (k/4) from the output of LocNet, and the bounding box is transferred back to the coordinate frame of the cropped volume. Then, the bounding box is applied to the cropped volume to obtain a smaller volume of size h × w × k, which is the target volume. One LocNet is trained for each target structure, which requires a bounding box of a specific size.

The second 3D U-Net, denoted as SegNet, is used to segment the target structure from the target volume obtained from the previous step. The target volume has a size of h × w × k, which is much smaller than the 384 × 384 × 224 cropped volume, and only one structure is segmented from it. These two characteristics make the segmentation of SegNet much easier. The output of SegNet is a mask volume with each voxel being 0 or 1, indicating background and target voxels, respectively.

LocNet and SegNet are separately trained; one LocNet and one SegNet are trained for each of the nine structures. In sections III-C and III-D, we introduce the preprocessing needed to prepare the training and testing data for the two 3D U-Nets and the concrete training and testing procedures.

### C. PREPROCESSING

1) INTERPOLATING AND CROPPING THE ORIGINAL IMAGES

The original images have different in-plane and inter-plane resolutions, which increase the variance of the shape and size of each structure and potentially increase the difficulty in segmenting them. Therefore, we resampled all the images into isotropic volumes with the same spatial resolution of 1 mm × 1 mm × 1 mm using bi-cubic interpolation. After interpolation, the in-plane size of all training and testing images was between 389 × 389 and 650 × 650 in voxels, and the number of slices was between 226 and 416.

Because the input size of SegNet and LocNet needs to be adjusted to multiples of eight, we need to crop the isotropic volumes after interpolation. Considering the sizes of the isotropic volumes in this dataset and the requirement that the size in each direction should be a multiple of eight, we cropped the images into a 384 × 384 × 224 volume automatically. However, we did not have to manually crop the training and testing images to place the target structures at the center of the cropped volume. In contrast, we divided the nine target structures into two groups and adopted a consistent cropping strategy for each group. The first group consisted of brainstem, optic chiasm, and optic nerves (both left and right), and the second group consisted of the mandible, parotid glands (both left and right), and submandibular glands (both left and right). The X, Y, and Z axes of the coordinate frame of the original images corresponded to the left-right, anterior-posterior, and superior-inferior directions of the human body. We positioned the 384 × 384 × 224 cropping window on the original images with margins on both sides of the cropping window along each axis. The voxels of the margins may be different for different images because of the differences in the size of images. Nevertheless, for all target structures, the ratio between the left and right margins along the X-axis was 0.5 to 0.5; the ratio between the anterior and posterior margins along the Y-axis was 0.3 to 0.7 and 0.2 to 0.8 for the structures in the first and second groups, respectively; the ratio between the superior and the inferior margins along the Z-axis was 0.9 to 0.1 and 0.7 to 0.3 for the structures in the first and second groups, respectively. Each target structure is not necessarily located at the center of the cropped volume because a dedicated network will be used to locate it. For the structures in each group, cropping is automatically performed on both the training and testing images with the same parameters.

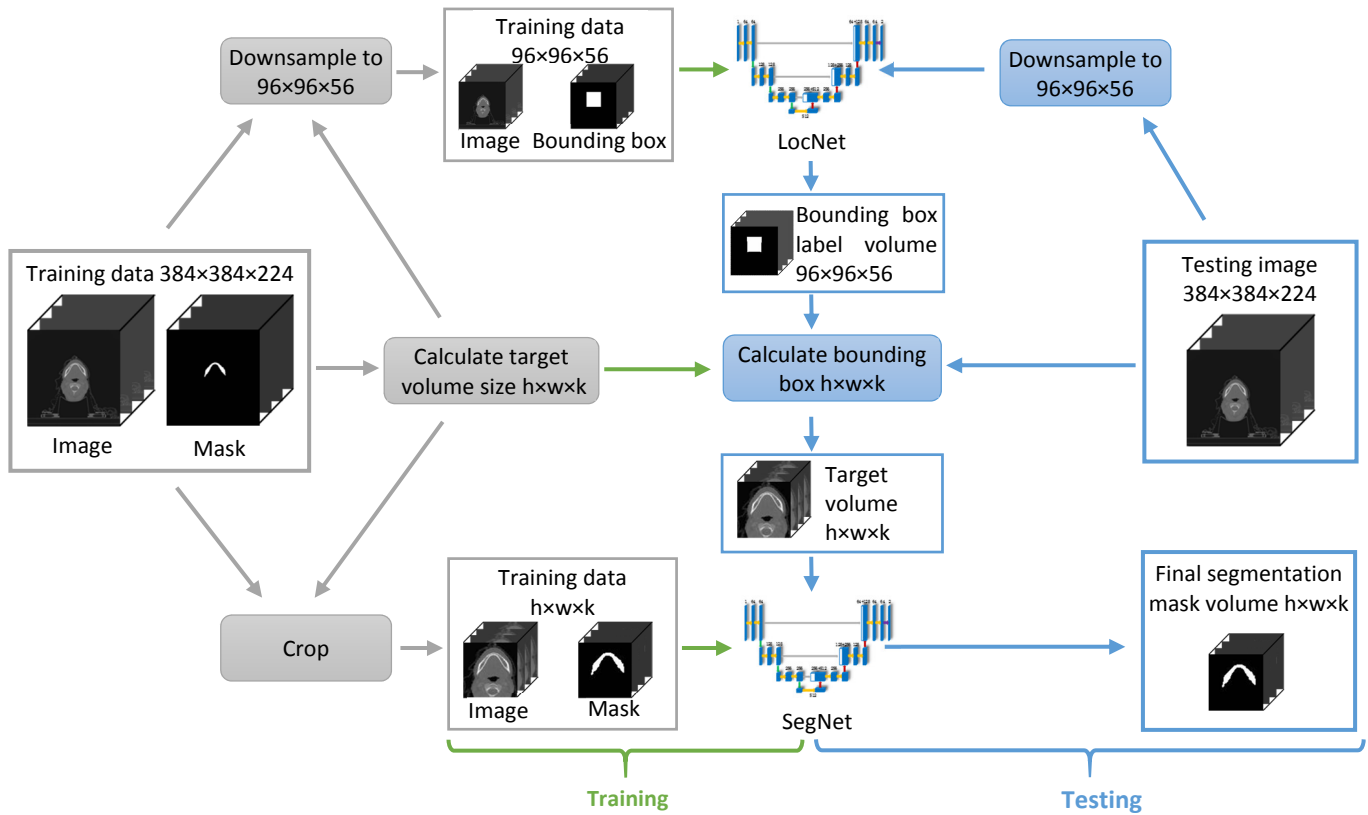

**FIGURE 2.** Flowchart of the proposed segmentation framework.

#### 2) DETERMINING THE SIZE OF THE BOUNDING BOX FOR EACH STRUCTURE

In the 384 × 384 × 224 cropped volume, we first located a bounding box to enclose the target structure and called the volume data within the bounding box as the target volume. We needed to determine the size of the bounding box for each structure before locating it. Because the target volume is the input of SegNet, its size in each direction should also be a multiple of eight. In this study, we determined the size of the target volume for each structure by considering the size of the structure in the training dataset (Table 1).

TABLE 1. Size of the bounding box for each target structure.

| Structure | Size |
| --- | --- |
| Mandible | 144 × 144 × 112 |
| Parotid left | 96 × 96 × 96 |
| Parotid right | 96 × 96 × 96 |
| Brainstem | 56 × 56 × 80 |
| Submandibular left | 48 × 48 × 64 |
| Submandibular right | 48 × 48 × 64 |
| Optic nerve left | 56 × 56 × 24 |
| Optic nerve right | 56 × 56 × 24 |
| Chiasm | 32 × 32 × 16 |

#### D. TWO-STAGE 3D U-NET SEGMENTATION FRAMEWORK

In this study, we concatenated two 3D U-Nets to segment a target structure, where the first 3D U-Net was used to locate a relatively small target volume that enclosed the target structure, and the second 3D U-Net was used to segment out the target structure from the target volume. The first and the second networks are called LocNet and SegNet, respectively. As shown in Fig. 3, LocNet and SegNet have the same network structure, consisting of an analysis path and a synthesis path. In the analysis path, each layer contains two 3 × 3 × 3 convolutions, each followed by a batch normalization (BN) and a rectified linear unit (ReLu), and then a 2 × 2 × 2 max pooling with strides of two in each dimension. In the synthesis path, each layer consists of an up-convolution of 2 × 2 × 2 by strides of two in each dimension, followed by two 3 × 3 × 3 convolutions each followed by a BN and a ReLu. Shortcut connections from layers of equal resolution in the analysis path provide essential high-resolution features for the synthesis path. In the final layer, a 1 × 1 × 1 convolution is used to reduce the number of output channels to a 0–1 classification. In total, each network has 17 convolutional layers.

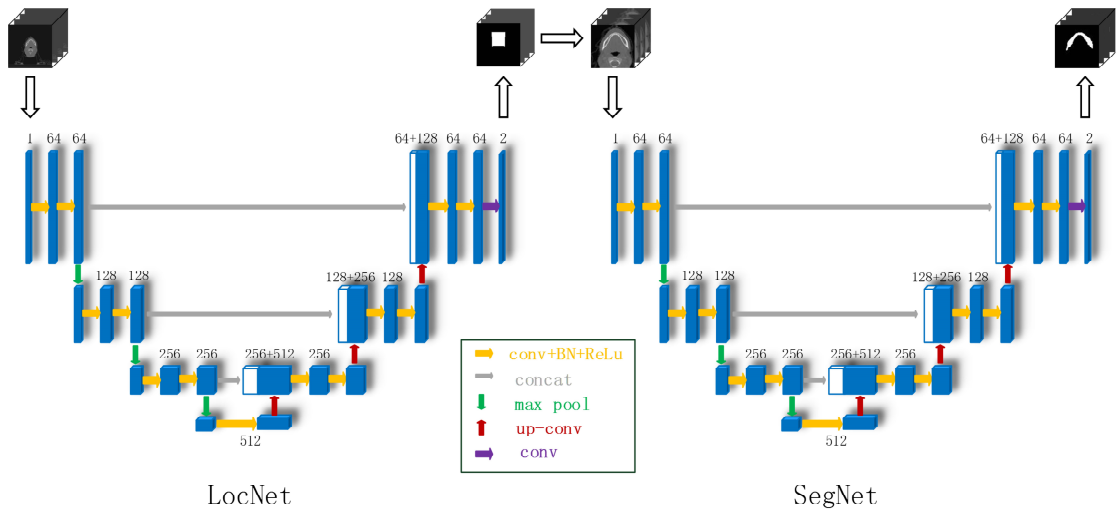

**FIGURE 3.** Structures of the two-stage 3D U-Net frameworks.

The 384 × 384 × 224 cropped volume is first downsampled to the size of 96 × 96 × 56 and then input into LocNet. The output of LocNet is a 96 × 96 × 56 sized binary volume, from which we locate the bounding box. The cropped volume is downsampled by a factor of four; thus, the bounding boxes that we want to locate in the 96 × 96 × 56 output volume also shrink by a factor of four. For example, the size of the bounding box of the mandible is 144 × 144 × 112 in voxels and thus we need to locate a bounding box with a size of 36 × 36 × 28 in the 96 × 96 × 56 output volume. It is very unlikely that all the voxels with a value of 1 fall in a cuboid of the expected size. Here, we used the sliding-window technique to locate the expected bounding box. We slid a cuboid with the expected size in the output volume and regarded the location at which the cuboid encloses the maximum number of voxels with value 1 as the true location of the bound box. When multiple locations have the same maximum number, the average location is used.

## III. EXPERIMENTS AND RESULTS

According to the regulation of the Head and Neck Auto-Segmentation Challenge 2015, we used all 33 training images in the dataset for training LocNet and SegNet and tested the segmentation framework using the 15 testing images. Four metrics were calculated to evaluate the performance of the proposed segmentation framework. We compared the proposed method with several state-of-the-art methods, including both traditional and artificial intelligence-based approaches. Finally, we showed the efficiency of the network by comparing the proposed method with two traditional approaches used in segmenting 3D medical images with a deep learning framework.

### A. EVALUATION METRICS
We used four evaluation metrics in this study.
1. Dice similarity coefficient (DSC). The DSC measures the degree of overlap between the segmentation result and the gold standard and is defined as follows:

$$\text{DSC} = \frac{2|A \cap B|}{|A| + |B|} \quad (1)$$

where $A$ and $B$ represent the voxel set of the segmentation result and the voxel set of the gold standard, respectively.

2. The 95% Hausdorff distance (95HD). Before defining 95HD, we first need to define the Hausdorff distance, which is usually used to measure the deviation of the contour of two areas. Given two-point sets $X$ and $Y$ and $d(x, y)$ measuring the Euler distance between the two points $x \in X$ and $y \in Y$, the directed Hausdorff distance can be defined as follows:

$$\overrightarrow{d_H}(X, Y) = \max_{x \in X} \min_{y \in Y} d(x, y) \quad (2)$$

Hausdorff distance $\overrightarrow{d_H}(X, Y)$ measures the largest distance from points in $X$ to its nearest neighbor in $Y$, and this distance is sensitive to large segmentation errors in a very small region. To eliminate this sensitivity, an 95% Hausdorff distance is calculated to measure the 95th percentile of the distance, denoted as $\overrightarrow{d_{H,95}}(X, Y)$. In this study, we used 95HD, which is calculated as follows:

$$\text{95HD} = (\overrightarrow{d_{H,95}}(X, Y) + \overrightarrow{d_{H,95}}(Y, X))/2 \quad (3)$$

3. Positive predictive value (PPV). The PPV is the proportion of the correctly segmented volume within the entire volume of segmentation result.

$$\text{PPV} = \frac{|A \cap B|}{|A|} \quad (4)$$

4. Sensitivity (SEN). SEN is the proportion of the correctly segmented volume within the entire volume of the gold standard.

$$\text{SEN} = \frac{|A \cap B|}{|B|} \quad (5)$$

### B. EXPERIMENTAL SETTINGS
The proposed networks were implemented using Python based on the Keras package [31] and experiments were performed on a computer with a single GPU (i.e., NVIDIA GTX 1080 Ti) and Linux Ubuntu 14.04 LTS 64-bit operating system.

We trained one LocNet and one SegNet for each of the nine OARs. The size of the training images for LocNet was 96 × 96 × 56 and the size of the training images for SegNet was determined by the size of the bounding box for each structure except for the mandible (Table 1). The size of the bounding box for the mandible was 144 × 144 × 112, but its target volume was further downsampled to 144 × 144 × 56 because of the memory limitation. The size of a mini-batch in each epoch was 1. Cross entropy loss function was adopted in the logistic regression as our loss function for both LocNet and SegNet, which was minimized by Adam optimizer using recommended parameters, and the training was terminated at 200 iterations over the training images. For each OAR, we trained one LocNet and one SegNet; thus, we trained 18 networks for all nine OARs in this dataset, which took approximately 30 hours. In testing stage, the segmentation of one OAR on one image took approximately 6 seconds, of which approximately 2 seconds was spent on the network processing of the image and approximately 4 seconds on the postprocessing of the output of LocNet. Several small isolated regions that did not belong to the target structure in the output of SegNet existed for some structures. We adopted a simple postprocessing technique, in which we deleted isolated regions whose volume was less than 10% of the total segmentation result.

*C. SEGMENTATION RESULTS OF THE PROPOSED METHOD*

We evaluated the performance of our method using the two-stage segmentation framework and interpolated isotropic images. In addition, we tested the proposed segmentation framework using the original images without interpolation. Without interpolation, the original images were not cropped; they were directly downsampled to a size of 128 × 128 × 64 as the input to LocNet. The size of the bounding box for some structures was different from that of the interpolated images, but the same size was used across all images. The processing after obtaining the target volume was the same as the interpolated images. In each experiment, DSC, 95HD, PPV, and SEN were calculated for each OAR and the results are listed in Table 2. In Table 2, the OARs are ordered by decreasing volume. The proposed method demonstrated good segmentation accuracy in large OARs and its performance decreased with the decrease in the volume of the OARs when the volume-related metrics, including DSC, PPV, and SEN, were considered. However, this observation did not hold for the contour-based metric, 95HD. Overall, the mean and standard deviation of the 95HDs for all the OARs were small, indicating that the segmentation method found the correct contour in most areas for each structure. The difference in the performance reflected by the volume-based and contour-based metrics is caused by the fact that similar levels of error on the contour can result in large errors in small structures and small errors in large structures when computing volume-based metrics because volume-based metrics use the volume of the structure as the denominator.

For most OARs, the segmentation results with interpolation were superior to those without interpolation, especially when the volume of the OAR was small. One possible reason for the decreased accuracy with interpolation is that the interpolated images have lower in-plane resolution than the original images. We interpolated the original images to 1 mm resolution in each dimension because of the memory limitation. Using a higher resolution for the interpolated images may further improve the accuracy of segmentation, not only for large OARs but also for small OARs.

Fig. 4 illustrates the segmentation results for subject 0522c0857 with and without interpolation. The mandible was segmented more accurately without interpolation, while the submandibular gland, optic nerve, and chiasm were segmented better with interpolation. To illustrate the overall performance of the proposed method, we show the segmentation results for subjects 0522c576, 0522c0667, and 0522c0857 with interpolation in Fig. 5. In addition, for each of the nine OARs, we chose one good segmentation result and one bad segmentation result and show the slices in Fig. 6. For the bad segmentation results, we thought it may be caused by the following three reasons. 1) Lack of training data. For we only have 33 training images, these were not enough for network training, and it was easy to make the network overfit. 2) Low contrast. The boundaries of some OARs are unclear, it is difficult to segment even for experienced radiologists, let alone neural networks. 3) Inaccurate localization. If the OAR can't be localized from the whole volume accurately by LocNet, it can't be segmented by SegNet accurately.

TABLE 2. AVERAGE (±STANDARD DEVIATION) PERFORMANCE OF OUR METHOD WITH AND WITHOUT INTERPOLATION (INT) FOR EACH STRUCTURE.

| Structure | DSC (%) | | 95HD (mm) | | PPV (%) | | SEN (%) | |
|---|---|---|---|---|---|---|---|---|
| | Without INT | With INT | Without INT | With INT | Without INT | With INT | Without INT | With INT |
| Mandible | **94.0±1.6** | 93.0±1.9 | **1.14±0.43** | 1.26±0.50 | 92.9±3.1 | **94.6±2.1** | **95.2±2.3** | 91.5±2.6 |
| Parotid left | 85.8±2.3 | **86.4±2.6** | 3.18±1.22 | **2.41±0.54** | 83.2±4.0 | **88.3±2.9** | **88.8±3.7** | 84.8±5.0 |
| Parotid right | **85.3±4.2** | 84.8±7.0 | 3.12±1.34 | **2.93±1.48** | **88.1±6.8** | 86.0±7.5 | 83.3±6.1 | **84.2±9.2** |
| Brainstem | 87.2±3.0 | **87.5±2.2** | 2.04±0.52 | **2.01±0.33** | **88.8±4.9** | 88.7±5.6 | 86.1±6.2 | **86.8±4.5** |
| Submandibular left | 73.8±10.5 | **75.8±14.7** | 3.89±2.54 | **2.86±1.60** | 72.5±15.0 | **82.2±10.4** | **78.9±15.6** | 74.4±17.7 |
| Submandibular right | 70.2±10.3 | **73.3±9.7** | 3.74±1.34 | **3.44±1.55** | 74.6±14.8 | **82.5±11.9** | 69.0±14.8 | **69.4±15.2** |
| Optic nerve left | 70.3±6.9 | **73.7±7.6** | **1.81±0.72** | 2.53±2.34 | 71.0±10.5 | **72.5±7.2** | 70.4±6.6 | **75.9±11.3** |
| Optic nerve right | 68.8±7.9 | **73.6±8.8** | 2.53±2.25 | **2.13±2.45** | 69.4±11.0 | **70.6±10.6** | 70.0±11.4 | **78.4±11.9** |
| Chiasm | 32.1±21.9 | **45.1±17.2** | 4.13±2.23 | **2.83±1.42** | 37.9±27.4 | **46.7±17.5** | 36.5±26.0 | **49.8±25.2** |

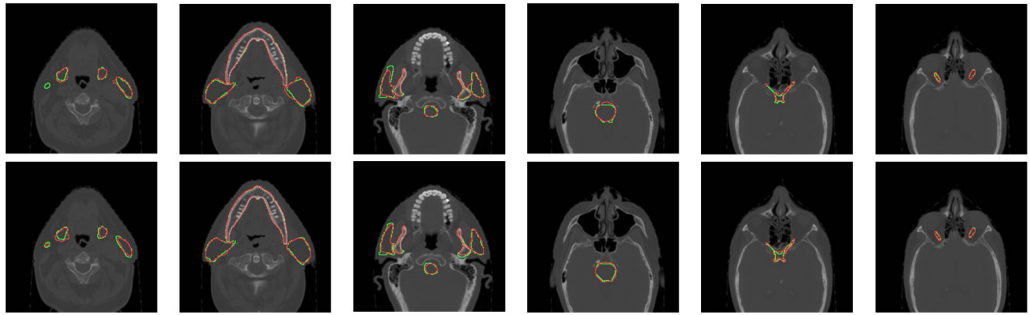

**FIGURE 4.** Segmentation results for subject 0522c0857. The first and second rows show the segmentation results without and with interpolation, respectively. From left to right: the 85th, 92th, 102th, 112th, 118th, and 120th slice of the axial view. The gold standard results are depicted in green, and our results are depicted in red.

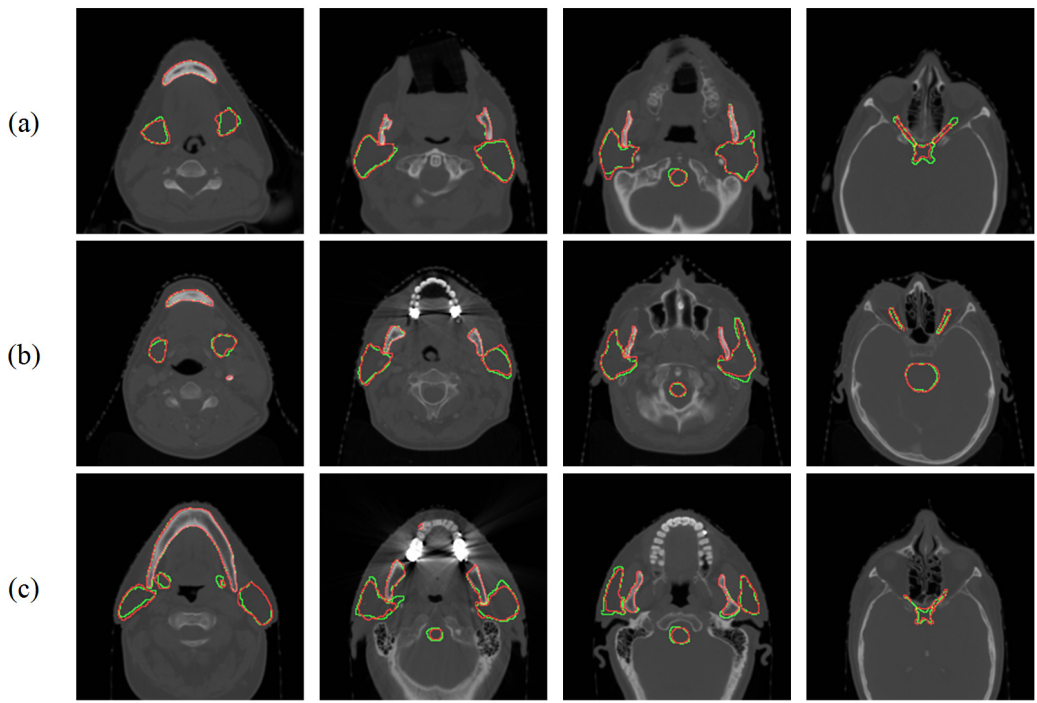

**FIGURE 5.** Segmentation results for subjects 0522c576 (row a), 0522c0667 (row b), and 0522c0857 (row c). The gold standard results are depicted in green, and our results are depicted in red.

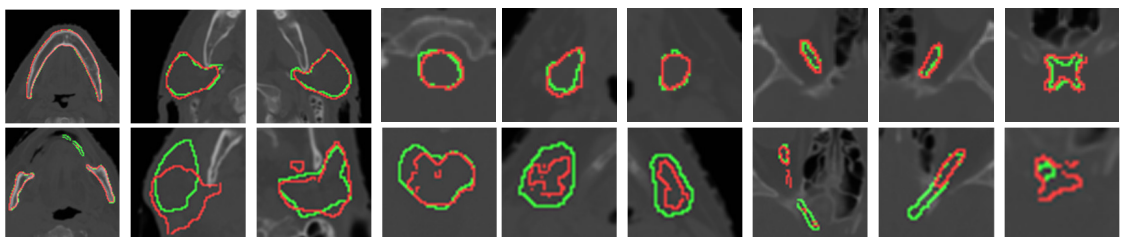

**FIGURE 6.** Example of typical good (upper row) and bad (lower row) segmentation results for the nine subjects (left to right: mandible, left parotid gland, right parotid gland, brainstem, left submandibular gland, right submandibular gland, left optic nerve, right optic nerve, and chiasm). The gold standard results are depicted in green, and our results are depicted in red.

### D. COMPARISON OF ACCURACY AGAINST STATE-OF-THE-ART METHODS

It is difficult to compare different methods of OAR segmentation in HaN CT images because of the differences in datasets, OARs, and evaluation metrics used in different studies. While MICCAI 2015 Challenge provides a unified evaluation framework. We first compare the proposed method (with interpolation) with the four methods that ranked top in the challenge. In these four methods, UC [32] provided DSCs for all nine OARs but provided no 95HDs, IM [33] provided DSCs and 95HDs for three OARs, and UB [11] and VU [34] provided DSCs and 95HDs for all nine OARs. Tables 3 and 4 illustrate the DSCs and 95HDs for our method and the four competing methods, respectively. As shown in Table 4, our method outperforms the competing methods in terms of 95HD with a large margin for seven of all nine OARs. In terms of the DSC, our method ranks first in five of the nine OARs and ranks second for the other three OARs.

TABLE 3. Average (± standard deviation) DSCs for the competing methods.

| Structure | Our method (with INT) | VU [34] | UC [32] | IM [33] | UB [11] | Tong [23] |
|---|---|---|---|---|---|---|
| Mandible | 93.0±1.9 | 91.7±2.4 | **93.1±2.4** | 91.0±2.0 | 88.0±6.2 | 87.0+3.0 |
| Parotid left | **86.4±2.6** | 81.1±4.0 | 76.6±7.2 | | 82.7±3.6 | 83.5±2.3 |
| Parotid right | **84.8±7.0** | 81.4±6.2 | 75.4±10.4 | | 80.5±6.6 | 83.2±1.5 |
| Brainstem | 87.5±2.2 | 80.3±6.5 | 85.6±4.8 | **88.0±3.0** | 84.9±4.7 | 87.0±3.0 |
| Submandibular left | **75.8±14.7** | 70.2±9.6 | 54.3±11.9 | | 72.3±7.3 | 75.5±6.5 |
| Submandibular right | 73.3±9.7 | 67.6±10.0 | 46.6±20.2 | | 72.3±10.5 | **81.3+6.5** |
| Optic nerve left | **73.7±7.6** | 64.4±7.1 | 48.2±15.0 | | 60.5±10.6 | 65.3±5.6 |
| Optic nerve right | **73.6±8.8** | 61.0±7.5 | 55.6±5.6 | | 63.9±9.0 | 68.9±4.5 |
| Chiasm | 45.1±17.2 | 38.0±15.7 | 8.1±8.0 | 38.0±2.0 | 55.7±12.1 | **58.4+10.3** |

TABLE 4. Average (± standard deviation) 95HDs for the competing methods (unit: mm).

| Structure | Our method (with INT) | VU [34] | IM [33] | UB [11] | Tong [23] |
|---|---|---|---|---|---|
| Mandible | **1.26±0.50** | 2.49±0.79 | 1.97 | 2.83±1.22 | 1.50+0.32 |
| Parotid left | **2.41±0.54** | 6.97±2.92 | | 5.11±1.25 | 3.97+2.15 |
| Parotid right | **2.93±1.48** | 6.43±2.56 | | 6.13±2.40 | 4.20+1.27 |
| Brainstem | **2.01±0.33** | 5.15±1.88 | 3.80 | 4.59±1.96 | 4.01+0.93 |
| Submandibular left | **2.86±1.60** | 5.35±1.75 | | 5.44±1.24 | 5.59+3.93 |
| Submandibular right | **3.44±1.55** | 5.74±2.57 | | 5.42±2.23 | 4.84+1.67 |
| Optic nerve left | 2.53±2.34 | 2.76±0.70 | | 3.23±0.99 | **2.52+1.04** |
| Optic nerve right | **2.13±2.45** | 3.15±1.27 | | 3.20±1.08 | 2.90+1.88 |
| Chiasm | 2.83±1.42 | 4.13±0.82 | 3.48 | 2.78±0.79 | **2.17+1.04** |

In addition to the four methods that can be directly compared, some other studies have used different datasets or the same dataset in a different training-testing grouping scheme.

The first study [20] using deep learning methods to segment OARs in HaN CT images provided DSCs for 13 OARs. Eight of these OARs were used in the MICCAI 2015 Challenge (except the brainstem). We cannot directly compare the results of our method with those of [20] because that study used a different set of data. Nevertheless, the DSC for our method is higher than that of [20] for seven OARs (6.0% higher on average). Additionally, this method needs a doctor to determine the approximate location of each OAR to be segmented. In [13], a hierarchical vertex regression-based segmentation method was proposed, and the DSCs for the brainstem, mandible, and parotid gland were 0.9±0.04, 0.94±0.01, and 0.84±0.06, respectively. However, this method was only evaluated by two-fold cross validation on 33 training images but not evaluated on the 15 testing images, and the segmentation results of other structures were not provided either. In [21], a interleaved 3D CNNs method was proposed to jointly segment the optic nerve and chiasm. The DSCs for the left optic nerve, right optic nerve, and chiasm were 0.72±0.08, 0.70±0.09, and 0.58±0.17, respectively. This method was designed to segment small targets and is not applied on other OARs in the MICCAI 2015 Challenge dataset. Furthermore, they utilized a joint segmentation scheme, while we segmented each OAR separately. In [22], the authors used an end-to-end and atlas-free three dimensional squeeze-and-excitation U-Net (3D SE U-Net) for fast and fully automated whole-volume HaN anatomical segmentation. The DSCs for the brainstem, chiasm, mandible, optic nerve left, optic nerve right, parotid left, parotid right, submandibular left and submandibular right are 0.867, 0.532, 0.925, 0.721, 0.706, 0.881, 0.874, 0.814, and 0.813, respectively. Though they used four datasets to train and test their model, our methods still get similar DSCs with their methods. We tried to use SE block in our SegNet, but it failed to impove the segmentation accuracy, and the results are listed in the supplemental file.

*E. COMPARISON OF RUNTIMES AGAINST STATE-OF-THE-ART METHODS*

Runtime comparison is difficult because the code of the competing methods is not available, and we cannot run all the methods on the same computer. Nevertheless, we listed the runtimes of VU, IM, UB, and Ibragimova given in the original papers for segmenting all nine OARs of one subject in Table 5. Segmenting all nine OARs using our method required approximately 108 s on average.

TABLE 5. Runtimes of the different methods.

| Method | Our method | VU | IM | UB | Ibragimova |
|---|---|---|---|---|---|
| Testing time (unit: min) | **1.8** | 20 | 30 | 12 | 4 |

*F. ROLE OF TARGET LOCALIZATION*

To show the superiority of the proposed target localization network, we compare our method with the following three baseline methods.

1) Joint Localization

In the localization stage, we trained one network for one structure, so we trained nine networks for nine structures. To demonstrate the superiority of one localization network for one structure, we trained a joint localization network for all nine structures. The joint localization network has the same structure as LocNet proposed in this paper, except that the output results are the location of nine structures. The output of joint localization network was processed by the same post-processing method as LocNet, and SegNet was used to segment each structure after joint localization.

TABLE 6. AVERAGE (±STANDARD DEVIATION) DSCs AND 95HDs FOR DIFFERENT STRATEGIES OF HANDLING LARGE VOLUMES.

| Method | Our method (with INT) | | Joint localization | | Downsampling | | Sliding-window | |
|---|---|---|---|---|---|---|---|---|
| | DSC (%) | 95HD (mm) | DSC (%) | 95HD (mm) | DSC (%) | 95HD (mm) | DSC (%) | 95HD (mm) |
| Mandible | **93.0±1.9** | **1.26±0.50** | 89.3±5.9 | 5.63±4.19 | 79.0±7.9 | 16.64±36.11 | 88.1±4.6 | 2.76±2.09 |
| Parotid left | **86.4±2.6** | **2.41±0.54** | 84.3±3.4 | 3.48±2.37 | 74.0±6.8 | 5.19±1.92 | 62.7±16.4 | 53.08±25.96 |
| Parotid right | **84.8±7.0** | **2.93±1.48** | 84.2±4.8 | 3.36±3.03 | 75.2±5.7 | 4.93±1.78 | 52.3±13.8 | 91.28±26.58 |
| Brainstem | **87.5±2.2** | **2.01±0.33** | 83.8±3.0 | 4.06±0.92 | 81.8±3.5 | 2.89±0.61 | 74.8±11.1 | 19.22±32.78 |
| Submandibular left | **75.8±14.7** | **2.86±1.60** | 69.8±4.7 | 3.32±1.82 | 65.5±5.7 | 3.72±1.08 | 66.3±9.0 | 57.89±46.67 |
| Submandibular right | **73.3±9.7** | **3.44±1.55** | 66.7±8.6 | 4.35±1.96 | 59.6±13.4 | 4.09±1.26 | 56.3±10.2 | 80.96±26.92 |
| Optic nerve left | **73.7±7.6** | **2.53±2.34** | 54.9±12.7 | 7.58±8.30 | 19.6±1.67 | 13.32±3.37 | 59.7±11.2 | 76.38±23.37 |
| Optic nerve right | **73.6±8.8** | **2.13±2.45** | 58.0±14.5 | 6.11±5.96 | 11.9±12.4 | 13.10±2.22 | 60.3±10.2 | 78.15±38.32 |
| Chiasm | **45.1±17.2** | **2.83±1.42** | 28.1±13.0 | 7.51±9.4 | 0 | infinity | 9.2±11.6 | infinity |

2) Downsampling

Because of the memory limit, it is challenging to place the entire 3D image volume into the GPU for training and testing. One solution is to downsample the original image to a manageable size. In the downsampling strategy used in this paper, we downsampled the training and testing images to a size of 96 × 96 × 56. Then we used nine 3D U-Net to segment nine target structure from the downsample image saperately.

3) Sliding-window

To solve the memory limit problem, another way adopted in many previous studies is sliding window [35], [36], which crops the original images into small blocks and performs the segmentation block by block. In the sliding-window strategy used in this paper, we cropped the original volume data to non-overlapping blocks with a size of 64 × 64 × 64. Of all the blocks, only a very small proportion contained the target structure, and thus we could not use all the blocks for training. Therefore, we kept all the blocks that contain the target structure and randomly chose the same number of blocks without the target structure for training SegNet. In the testing stage, we slid a window of size 64 × 64 × 64 in the whole volume with some overlap between neighboring windows and adopted a maximum voting for each voxel to obtain the final segmentation result.

The DSCs and 95HDs for the three baselines and the proposed method with interpolation are listed in Table 6. As shown in Table 6, our method outperforms the other methods in both DSC and 95HD. For joint localization method, though it reduced the number of LocNets, it needed the network focus on nine different structure localization, which was difficult for only one model. Due to the location error of joint localization method, the segmentation accuracy of target structure is slightly lower than that of the proposed method. For downsampling method, it is very difficult to distinguish small structures, such as the optic nerve and chiasm, in the downsampled images, and thus their segmentation accuracy was very low. For the sliding-window method, several parameters, such as the window size, step size, and ratio between the positive and negative samples for training, may influence the final result. We experimented with several combinations of parameters and kept the best one, but we cannot guarantee that the reported accuracy is the best possible result. The 95HDs for the sliding-window strategy were very large for most OARs because this method segments out some false-positive voxels far from the true target OAR. Some postprocessing strategies may improve the accuracy of these two strategies, but the improvement is limited.

As shown in Table 7, we compared the average training and testing time of four methods for segmenting one structure. For the training time, our method needs to train two networks for each structure, so it takes longer than joint localization and downsampling. For the testing time, our method needs to locate and segment one structure using two networks, between which some processing needs to be done to obtain the bounding box. Therefore, the testing time of the proposed method is longer than that of downsampling. The training and testing time of the sliding-window strategy are much longer than those of the proposed method.

TABLE 7. RUNTIMES OF THE PROPOSED METHOD AND DOWNSAMPLING AND SLIDING-WINDOW STRATEGIES.

| Method | Proposed method | Joint localization | Downsampling | Sliding-window |
|---|---|---|---|---|
| Training time (unit: hour) | 3.2 | **1.4** | 1.8 | 6.3 |
| Testing time (unit: second) | 4.0 | 4.0 | **0.9** | 40 |

## IV. DISCUSSION

In this study, we proposed a new framework for the automatic segmentation of OARs in HaN CT images and evaluated its performance with the MICCAI 2015 Challenge dataset. In contrast to the previous methods that are based on deep neural networks, the proposed framework decomposes the segmentation into two simpler tasks: locating a bounding box and segmenting a small volume within the bounding box and trains a 3D U-Net for each task. The proposed two-stage framework easily achieves a large field of view with a small memory print. If a small structure is going to be segmented in a large image, 3D U-Net can produce positive values at irrelevant parts of the large image, since the field of view will be too small, which prevents it from understanding that the given location is irrelevant. Using a LocNet provides this large field of view with a small memory footprint. Experiments using the MICCAI 2015 Challenge dataset showed that the proposed method significantly outperformed the state-of-the-art methods.

There are many methods that have been used for the segmentation of OARs in HaN CT images, but the results were relatively poor compared to other medical image segmentation tasks. The difficulty comes from the characteristics of the OAR segmentation task, such as the large variability in the shape and size across different target structures and the poor contrast between some structures and their background. Deep neural networks have become the best choice for most image processing tasks and often outperform traditional methods with a large margin in medical image segmentation applications [14]–[19]. However, existing studies on the application of deep neural networks in OAR segmentation in HaN CT images demonstrate similar performance to traditional methods. One of the major obstacles in using deep neural networks in medical image segmentation has been the contradiction between large-sized high-resolution images and limited memory. Previously, this problem was addressed by using downsampling or sliding-window strategies, but our experiments show that the performance of both strategies is very poor. In a recent study, a multiatlas-based segmentation method [21] was first used to roughly locate the region of interest and then segmented only a small volume within the region using a 3D CNN. This method achieved high segmentation accuracy on three small structures and showed that decomposing the localization and segmentation tasks is helpful.

In this study, we utilized 3D U-Net for both localization and segmentation tasks. The decomposition made each of the two tasks much easier, and the deep neural network could be properly trained for the specific task. The results showed that the trained LocNet could locate the bounding box containing the target structure in all cases. After the bounding box was accurately located, training the SegNet to segment one structure with a similar shape and appearance in different subjects became much easier than training a network to segment multiple structures with different shapes and appearances from the original images. This strategy of decomposing a medical image segmentation task into two tasks, i.e., locating a bounding box and segmenting in the bounding box, has also been used in other applications where multiple structures are to be segmented.

In this study, 3D U-Nets were used for both the locating and segmentation tasks, and many other network structures can be used to replace the 3D U-Net for one task or both tasks. We did not attempt to test different network structures in this study, but experimenting with more network architectures is a potential research direction in the future. For some of the output of SegNet, there were several small isolated regions that did not belong to the target structure. The simple postprocessing adopted in this study only slightly improved the final results, and a more sophisticated postprocessing method may further improve the accuracy. In addition, the number of subjects in the MICCAI 2015 Challenge dataset is not very large, which may limit the performance of the deep learning network. In the future, verifying whether the segmentation accuracy of the method can be further improved by training on more data is necessary. Moreover, testing whether the method is suitable for clinical use and whether it can help improve treatment planning workflows are important.

## V. CONCLUSION

In this study, we proposed a two-stage segmentation framework based on 3D U-Net for the automatic segmentation of OARs in HaN CT images. The framework decomposes the original segmentation tasks into two easier subtasks: locating a bound box of the target structure and segmenting the target structure in a small volume within the bounding box. One 3D U-Net is trained for each task, and the decomposition allows the two tasks to be completed more accurately and quickly. Experiments using the MICCAI 2015 Challenge dataset show that the proposed method significantly outperforms the state-of-the-art methods.

## ACKNOWLEDGMENT
Yueyue Wang and Liang Zhao are co-first authors.

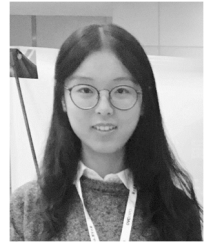

**Yueyue Wang** received the bachelor's degree in communication engineering from Ocean University of China, in 2017. She is currently pursuing the Ph.D. degree with the Digital Medical Research Center, School of Basic Medical Sciences, Fudan University, and the Shanghai Key Laboratory of Medical Imaging Computing and Computer Assisted Intervention, Shanghai, China. Her research interest medical image segmentation and medical image classification.

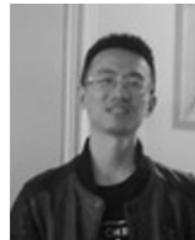

**Liang Zhao** received the bachelor's degree in physics from Nankai University, Tianjing, China, in 2005, the MSc in medical physics from the University of surrey, UK, in 2007. He is currently pursuing the Ph.D degree with the Digital Medical Research Center, School of Basic Medical Sciences, Fudan University, and the Shanghai Key Laboratory of Medical Imaging Computing and Computer Assisted Intervention, Shanghai, China. His research interest includes medical image processing/analysis and edge computing.

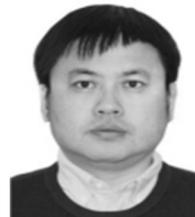

**Manning Wang** received the B.S. and M.S. degrees in power electronics and power transmission from Shanghai Jiaotong University, Shanghai, China, in 1999 and 2002, respectively, and the Ph.D. degree in biomedical engineering from Fudan University, Shanghai, in 2011. He is currently a Professor of biomedical engineering with the School of Basic Medical Science, Fudan University, where he is also the Deputy Director of the Digital Medical Research Center and the Shanghai Key Laboratory of Medical Imaging Computing and Computer Assisted Intervention (MICCAI). His research interests include medical image processing, image-guided intervention, and computer vision.

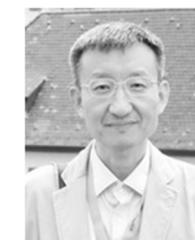

**Zhijian Song** received the B.S. degree from the Shandong University of Technology, Shandong, China, in 1982, the M.S. degree from the Jiangsu University of Technology, Jiangsu, China, in 1991, and the Ph.D. degree in biomedical engineering from Xi'an Jiaotong University, Xi'an, China, in 1994. He is currently a Professor with the School of Basic Medical Science, Fudan University, Shanghai, where he is also the Director of the Digital Medical Research Center and the Shanghai Key Laboratory of Medical Image Computing and Computer Assisted Intervention (MICCAI). His research interests include medical image processing, image-guided intervention, and the application of virtual and augmented reality technologies in medicine.